%% file: samplepaper.tex
\documentclass[runningheads]{llncs}
\usepackage[T1]{fontenc}
\usepackage{graphicx}
\usepackage{amsmath}
\usepackage{amssymb}
\usepackage{graphicx}
\usepackage{wrapfig}
\usepackage{float}
\usepackage{multirow}
\usepackage[table]{xcolor}
\usepackage{booktabs}
\usepackage{subcaption}
\usepackage{marvosym}
\usepackage[colorlinks=true, citecolor=blue]{hyperref} 
\definecolor{defaultcolor}{HTML}{E8E2F7}

\usepackage[numbers]{natbib} 
\usepackage{etoolbox}

\makeatletter
\patchcmd{\thebibliography}
  {\bibsection}
  {\bibsection\vspace{-2.5\baselineskip}} 
  {}{}
\makeatother

\input{commands}
\title{Inference-Time Scaling for Visual AutoRegressive modeling by Searching Representative Samples}
\author{Weidong Tang \and Xinyan Wan \and Siyu Li \and Xiumei Wang\textsuperscript{(\Letter)}}
\authorrunning{W.Tang et al.}
\institute{
School of Electronic Engineering, Xidian University, Xi'an, China\\
\email{	wangxm@xidian.edu.cn} 
}

\begin{document}
%
%
%
%
%
\maketitle              
\vspace{-20pt}
\begin{abstract}

While inference-time scaling has significantly enhanced generative quality in large language and diffusion models, its application to vector-quantized (VQ) visual autoregressive modeling (VAR) remains unexplored. We introduce \textbf{VAR-Scaling}, the first general framework for inference-time scaling in VAR, addressing the critical challenge of discrete latent spaces that prohibit continuous path search.
We find that VAR scales exhibit two distinct pattern types: general patterns and specific patterns, where later-stage specific patterns conditionally optimize early-stage general patterns.
To overcome the discrete latent space barrier in VQ models, we map sampling spaces to quasi-continuous feature spaces via kernel density estimation (KDE), where high-density samples approximate stable, high-quality solutions.
This transformation enables effective navigation of sampling distributions. We propose a density-adaptive hybrid sampling strategy: Top-$k$ sampling focuses on high-density regions to preserve quality near distribution modes, while Random-$k$ sampling explores low-density areas to maintain diversity and prevent premature convergence.
Consequently, VAR-Scaling optimizes sample fidelity at critical scales to enhance output quality. Experiments in class-conditional and text-to-image evaluations demonstrate significant improvements in inference process. The code is available at \url{https://github.com/WD7ang/VAR-Scaling}.

\end{abstract}

\vspace{-30pt}
\section{Introduction}
\vspace{-10pt}

Recent research has yielded significant advances in inference-time scaling laws for generative models~\cite{pope2023efficiently,leviathan2023fast,wu2025inference}. Large language models (LLMs) demonstrate powerful implicit chain-of-thought (CoT) capabilities by emulating deliberate, System 2-like reasoning~\cite{kojima2022large,fei2023reasoning,wei2022chain,yu2023towards}, internally exploring multi-step, multi-path solutions to select optimal outputs. Concurrently, a new scaling paradigm has emerged in diffusion models~\cite{peebles2023scalable}: instead of increasing denoising steps, researchers now use verifiers with intelligent search algorithms to actively explore noise space for superior paths~\cite{ma2025inference}. This approach strategically leverages extra computation via directed search, significantly boosting the quality of output.

\begin{figure*}[t]
    \centering
    \includegraphics[width=1\textwidth]{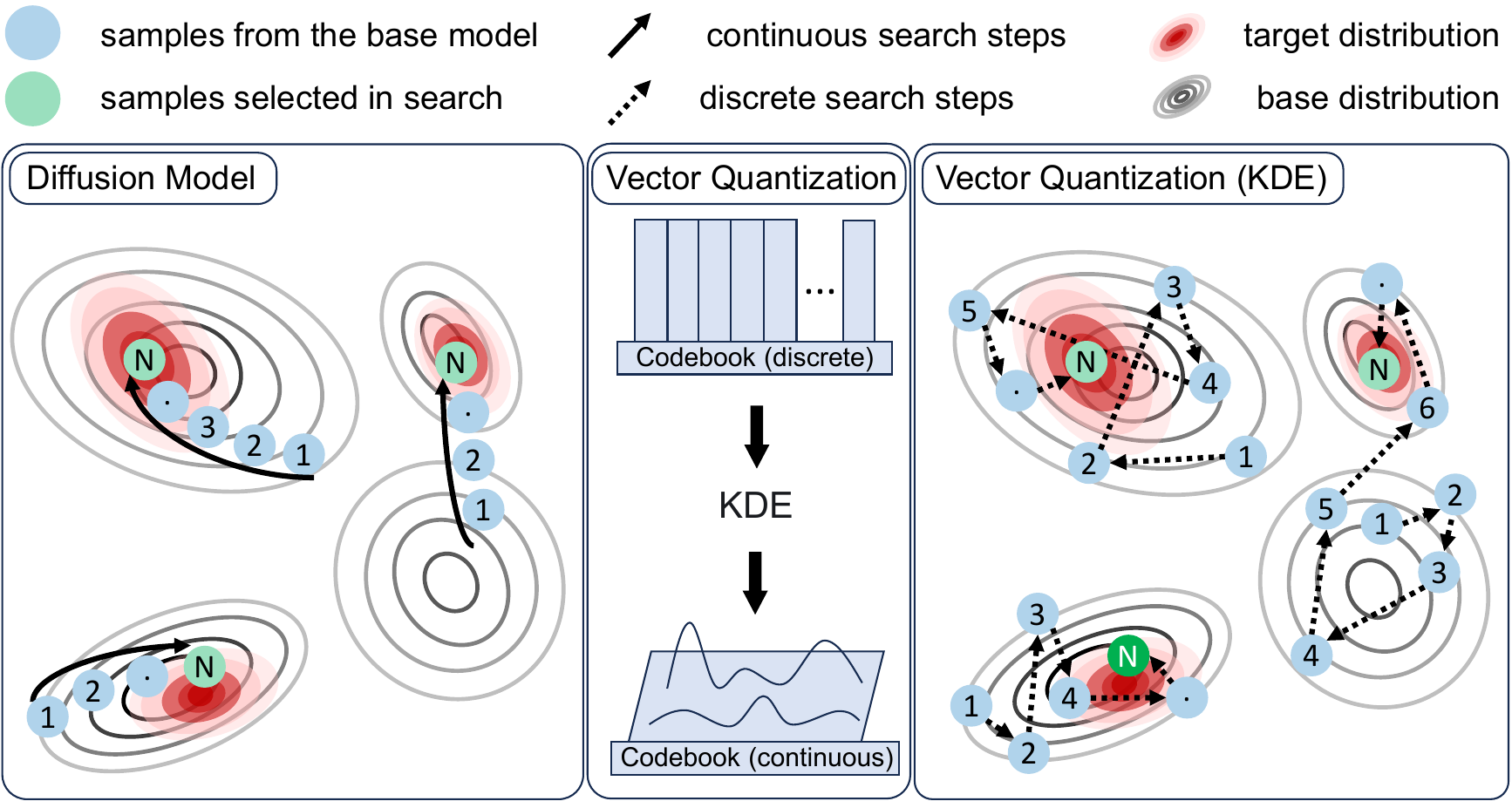}
    \caption{Diffusion models enable metric-guided continuous search from base to target distributions. In contrast, VQ's discrete nature prevents continuous path exploration. Our KDE mapping transforms probability-scaled sampling spaces into continuous trajectories, eliminating chaotic unguided searches evident.}
    \label{fig:problem}
    \vspace{-15pt}
\end{figure*}

While inference-time scaling uses LLMs and diffusion models, it remains unexplored in VQ generative models. 
Currently, the state-of-the-art VQ framework, VAR achieves superior fidelity through its “next-scale prediction” paradigm~\cite{tian2024visual}.
Therefore we adopt VAR~\cite{tian2024visual} as the baseline to evaluate novel scaling strategies for discrete generative spaces. 
Crucially, as VAR operates through a coarse-to-fine generation process, the quality of final output is primarily determined by the quality of coarse scale sampling decisions~\cite{tian2024visual}. 
In the VAR inference process, We find that: \emph{i)} General patterns, \textit{e.g.}, categories, object outlines, are identified in the beginning inference iterations; \emph{ii)} Specific patterns, \textit{e.g.}, texture details, edge refinements, are incrementally incorporated in the later inference iterations\footnote{Detailed implementations about these vanilla methods are in Sec.~\ref{subsec:perturbation}}. 
Consequently, improving coarse-scale sampling quality directly minimizes initial error sources, thereby mitigating downstream propagation effects.

Therefore, to enhance sampling quality at specific coarse scales, we must address two fundamental challenges: \emph{i)} defining what constitutes representative samples, and \emph{ii)} developing efficient methods to identify them. As shown in Figure~\ref{fig:problem}, while diffusion models locate high quality samples through noise space search guided by verifier scores~\cite{qi2024not,thaker2024frequency}, the discrete latent space of VQ models impedes direct transfer of such approaches. Consequently, we conduct dedicated analysis for discrete spaces by implementing two baseline strategies: random-k, top-k and small-k sampling to evaluate how sampling strategies impact VAR's performance across different sample sizes. Through empirical observations of VAR's inference process, we find that repeated sampling from the predictive distribution enables KDE-based Euclidean mapping. This transforms discrete sampling spaces into quasi-continuous feature spaces. Within these spaces, high-density samples represent the most characteristic outputs\footnote{Detailed implementations about these vanilla methods are in Sec.~\ref{subsec:inference_scaling}}.

Therefore, we propose \textbf{VAR-Scaling}, the first framework for inference time scaling in VAR. Our approach progressively clarifies representative samples at specific coarse scales of VAR by expanding the sample space, while strategically preserving diversity through density-adaptive sampling: when high-density regions are detected in the current distribution, we employ top-k sampling; otherwise, random-k sampling is activated. This architecture unlocks VAR's latent potential without exhaustive discrete space search, achieving significant performance gains through intelligent sampling allocation.

Our method demonstrates effective performance in both class-conditional and text-conditional image generation. We conduct comparative testing against VAR and its relevant works. The contributions are as follows:
\vspace{-5pt}
\begin{enumerate}
    \item We propose VAR-Scaling, the first general framework for inference-time scaling in VAR. We show that searching representative samples through scaling number of samples can lead to improvement generation tasks.
    \item Based on experiments, we approximately map the discrete space to the continuous space, and empirically define the sample with the highest density as the representative sample. It provides a new idea for the inference time scaling of discrete Spaces for the first time. 
    \item The evaluation results on VAR, FlexVAR~\cite{jiao2025flexvar}, and Infinity~\cite{han2025infinity} demonstrate that VAR and FlexVAR improve the IS~\cite{salimans2016improved} score by 8.7\% and 6.3\% respectively while maintaining stable FID~\cite{heusel2017gans}.  On Infinity, the Geneval score increases 1.1\%.
\end{enumerate}

\vspace{-15pt}
\section{Related Work}
\vspace{-7pt}
\subsection{Inference-time scaling}
\vspace{-5pt}
In recent years, inference-time scaling for LLMs has seen remarkable progress~\cite{borzunov2023distributed,geiping2025scaling,wu2025inference,snell2024scaling}. Approaches that integrate search algorithms with LLMs have emerged, such as breadth - first search~\cite{wang2025planning}, depth - first search~\cite{tarjan1972depth}, Monte Carlo Tree Search~\cite{browne2012survey}, and guided beam search~\cite{pryzant2023automatic,xie2023self}. These methods highlight the benefits of incorporating search during inference, as they can boost performance across various tasks. In addition, process reward models (PRM)~\cite{yao2023tree} have proven effective in selecting reasoning paths with low error rates and providing rewards within reinforcement learning - style algorithms~\cite{lee2024rlaif,ma2023lets,wangself}. By rewarding intermediate steps, PRM can guide multi - step reasoning processes~\cite{lightman2023let}. However, the application of these approaches in the field of image generation has been limited. While some recent research has focused on iterative processes in diffusion models~\cite{ho2020denoising,song2021denoising,dhariwal2021diffusion} and the use of Chain of Thought (CoT) for self - correction in autoregressive image generation~\cite{lee2022autoregressive,esser2021imagebart}, visual autoregressive models based on VQ models have not been thoroughly investigated.
\vspace{-7pt}
\subsection{Autoregressive Visual Generation}
\vspace{-5pt}
The visual autoregressive models based on VQ use vector quantization to convert image patches into index tokens. These tokens are predicted at a coarse scale and then used to predict and fuse the residual image patches at the next scale, gradually generating a finer-resolution image~\cite{tian2024visual,lee2022autoregressive}. However, due to the limitations of quantization errors, this approach struggles to further improve performance, especially in generating high-resolution images where details are often compromised~\cite{esser2021taming,van2017neural}. To address this issue, the VAR-based Infinity model employs an almost infinitely large continuous tokenizer, bypassing the limitations of discrete quantization and significantly enhancing generation quality~\cite{yao2023tree}. By using continuous tokens, the model is able to handle image generation more precisely, improving the depiction of finer details.

\vspace{-5pt}
\section{Methodology}
\vspace{-5pt}
In this Section, we first introduce VAR~\cite{tian2024visual} and KDE in Section~\ref{VAR}. 
Sec.~\ref{subsec:perturbation} establishes coarse-scale sampling quality as the fidelity determinant in VAR. 
Three inference search strategies reveal sample distribution properties, yielding our density-guided algorithm from empirical data.

\subsection{Preliminary}

\subsubsection{Generative Priors in VAR.}
\label{VAR}
Different from traditional autoregressive methods, VAR redefines image generation as a coarse-to-fine task. In VAR, the autoregressive unit is an entire token map, rather than a single token. It start by quantizing a feature map $f \in \mathbb{R}^{h\times w\times C}$into $K$ multi-scale token maps $(r_1, r_2, \dots, r_K)$, each at a increasingly  higher resolution $h_k\times w_k$, culminating in $r_K$ matches the original feature map's resolution $h\times w$.
The autoregressive likelihood is formulated as:
\begin{align}
    p(r_1, r_2, \dots, r_K) = \prod_{k=1}^{K} p(r_k \mid r_1, r_2, \dots, r_{k-1}),  \label{eq:var}
\end{align}
where each autoregressive unit $r_k \in [V]^{h_k \times w_k}$ is the token map at scale $k$ containing $h_k \times w_k$ tokens, and the sequence $(r_1, r_2, \dots, r_{k-1})$ serves as the the ``prefix'' for $r_k$.

\vspace{-5pt}
\subsubsection{Density estimation.}
KDE is a non-parametric method for estimating unknown probability density functions from finite samples. Given $n$ candidate tokens $\{x_i\}_{i=1}^n \subset \mathbb{R}^d$ sampled from a discrete codebook, the multivariate density estimator with isotropic Gaussian kernel~\cite{scott1992multivariate} is defined as:
\begin{equation}
\hat{f}(x) = \frac{1}{n(2\pi)^{d/2}h^d} \sum_{i=1}^n \exp\left(-\frac{\|x - x_i\|^2}{2h^2}\right)
\end{equation}
where $h > 0$ is the bandwidth parameter controlling smoothness, and $d$ denotes the embedding dimension. The bandwidth selection follows Silverman's rule generalized for multivariate data:
\begin{equation}
h = \sigma \left[\frac{n(d+2)}{4}\right]^{-1/(d+4)}
\end{equation}
where $\sigma$ is the dimension-wise mean standard deviation $\frac{1}{d}\sum \sigma_j$. The exponential term $\exp(-\|x - x_i\|^2/(2h^2))$ measures similarity between tokens, with the summation reflecting density accumulation from neighboring samples.

\subsection{The inference process of VAR.}
\label{subsec:perturbation}
VAR employs a progressive refinement strategy across multi-scale stages $k \in \{0,1,2,\dots,9\}$. At scale $k$, the model decodes a multi-scale token map $r_k$ (resolution $h_k \times w_k$) to generate output images:
\begin{equation}
I_k = D(r_k),
\end{equation}
where $I_k$ denotes the generated image at scale $k$, and $D$ represents the decoder function. The process initiates from a low resolution base ($k=1$) and incrementally adds details. 
To investigate the boundary between universal and specific patterns, we substituted the random number generator at scale $k$ and observed resultant variations in the final decoding output.

\begin{minipage}{0.35\textwidth}
\vspace{8pt}
\setlength{\leftskip}{-0.5cm}
\subsubsection{Analysis of Perturbation Results.}
Figure~\ref{fig:var_scale} shows that perturbing at scale 1 leads to significant divergence, highlighting its critical role in defining the final image's spatial structure. At scale 2, differences are noticeably smaller. Subsequent scales refine initial selections with perceptually nuanced variants, incorporating fine details into established general patterns, enhancing overall image quality.
\end{minipage}%
\hfill
\begin{minipage}{0.58\textwidth}
\centering
\vspace{10pt}
\includegraphics[width=\linewidth]{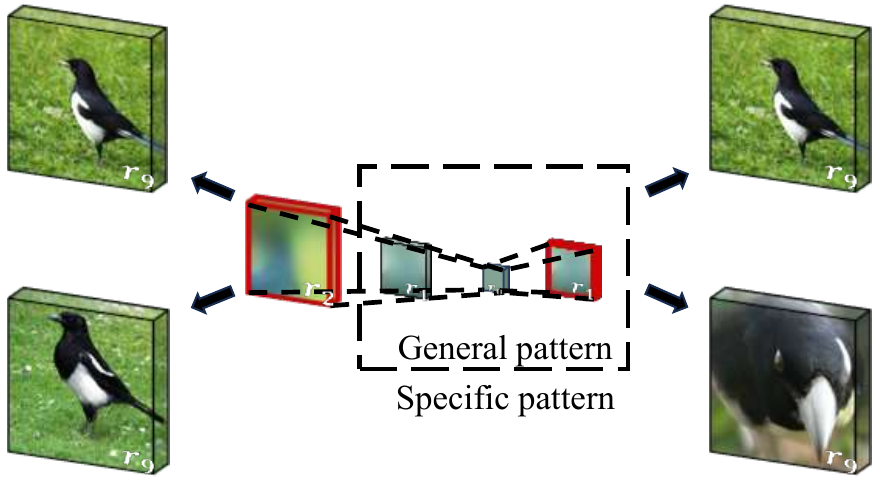}
\captionof{figure}{Scales 0-1 define general patterns like spatial structures and contours; scales 2-9 refine specific patterns with texture, edge, and local feature enhancements.}
\label{fig:var_scale}
\end{minipage}

\subsection{Inference-time Scaling on VAR.}
\label{subsec:inference_scaling}
To identify representative samples, we first assess inter-sample distances via Euclidean metric, then compute density scores using KDE. Three baseline selection strategies are proposed:

\begin{figure*}[h]
    \centering
    \begin{subfigure}[b]{0.48\textwidth}
        \centering
        \includegraphics[width=\linewidth]{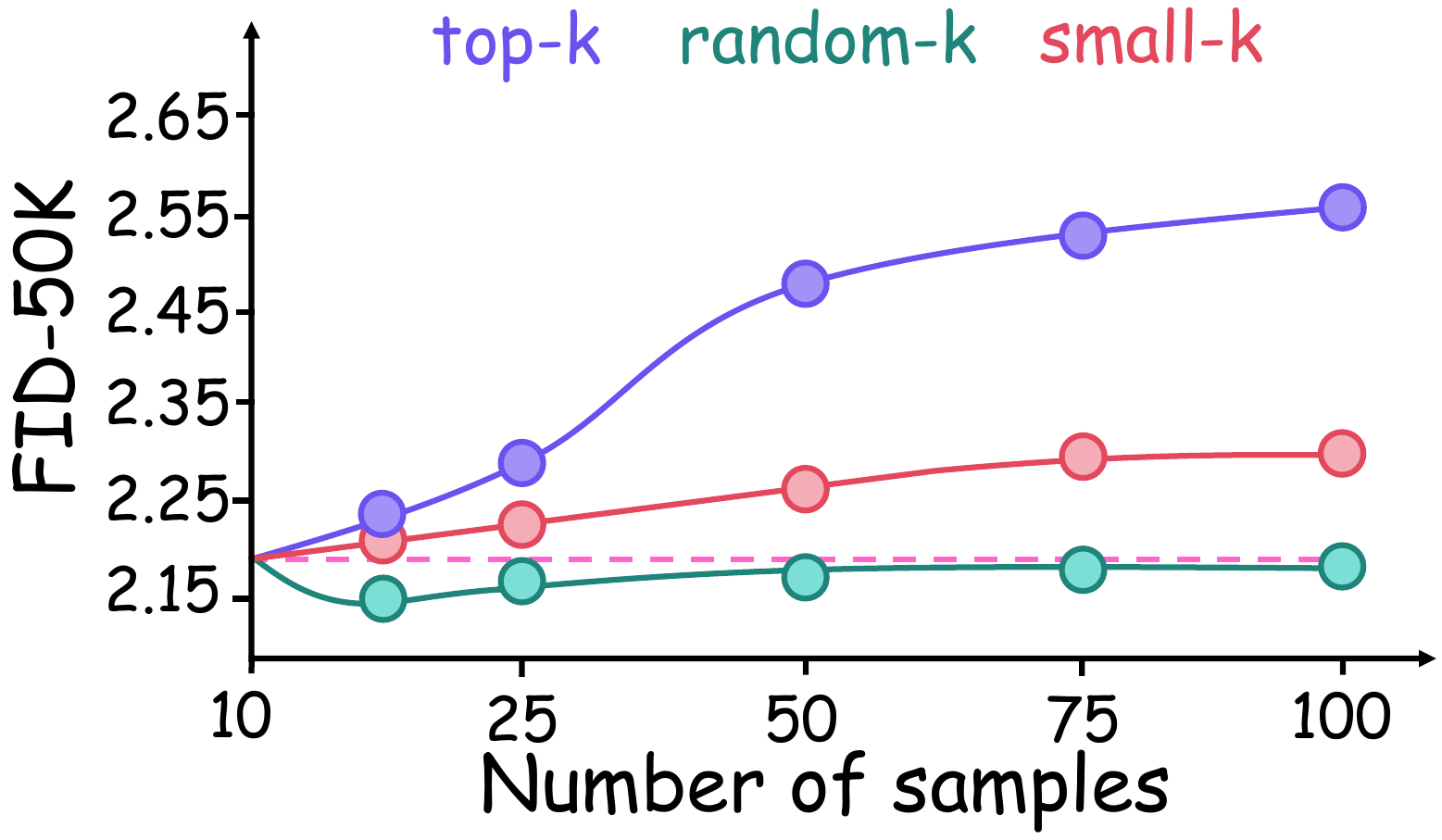}
        \caption{FID evaluated on ImageNet-50k}
        \label{fig:fid}
    \end{subfigure}
    \hfill
    \begin{subfigure}[b]{0.48\textwidth}
        \centering
        \includegraphics[width=\linewidth]{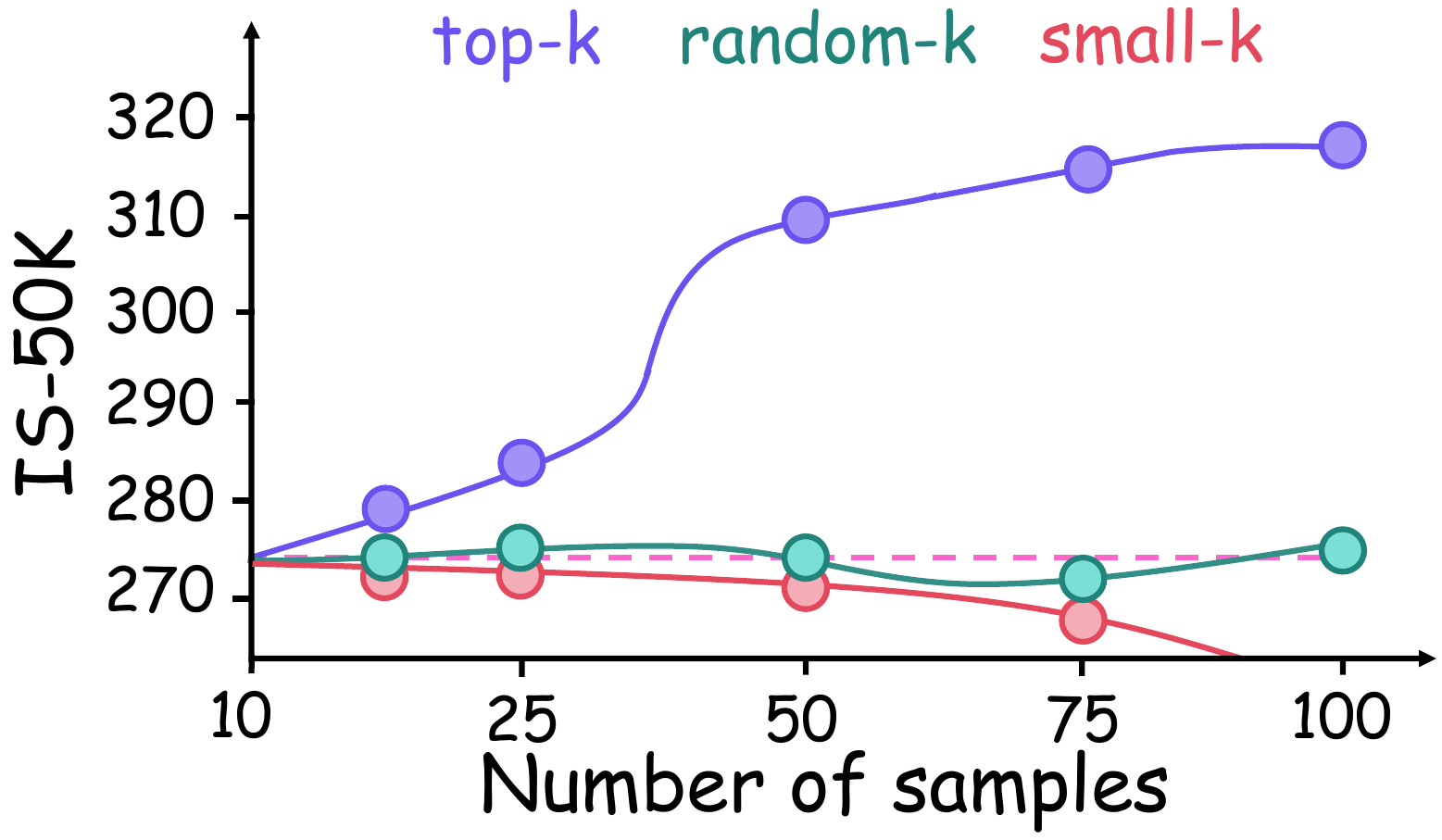}
        \caption{IS evaluated on ImageNet-50k}
        \label{fig:is}
    \end{subfigure}
    \caption{For ImageNet generation, we report model performance using the standard evaluation metrics, FID $\downarrow$~\cite{heusel2017gans} and IS $\uparrow$~\cite{salimans2016improved}. Our experiments show that high-density samples are likely representative prototypes, while low-density samples may correspond to low quality outliers.}
    \label{fig:top_random_small}

\end{figure*}

\subsubsection{Top-$k$ Sampling.} 
Density-guided sampling selects representative samples that balance diversity and quality, selecting the top k highest-density samples and using their densities as selection probabilities.

\begin{equation}
S_{\text{top-}k} = \{ x_i \mid x_i \in \mathop{\arg\max}\limits_{x \in \mathcal{X}} \hat{f}(x),\ i = 1, 2, \ldots, k \},
\end{equation}
\begin{equation}    
\mathrm{Top}_{k}(m, \mathcal{X}) :=
\left\{
\begin{aligned}
\mathop{\arg\max}\limits_{|\kappa|=k,\kappa\subset\mathcal{X}} \sum_{\kappa_{j}\in\kappa}m(\kappa_{j}),& \text{~if~} |\mathcal{X}| > k, \\
\mathcal{X},& \text{~otherwise.}
\end{aligned}
\right.
\end{equation}
where $\mathcal{X}$ is the sampling space and $\hat{f}(x)$ is the density score of sample $x$. The probability of selecting each sample $x_i$ is normalized.

However, as shown in Figure~\ref{fig:top_random_small}, with inference time scaling, although IS increases significantly, FID rises uncontrollably. This is because as the sampling space expands, the distribution characteristics of samples become more apparent. This biased search leads to multiple samplings of certain representative samples, causing low variance and high similarity among samples. This guided search approach shares conceptual similarities with \emph{reward hacking} in reinforcement learning, and existing research terms this phenomenon as \emph{verifier hacking}~\cite{ma2025inference}.

\subsubsection{Small-$k$ Sampling.}
As a conceptual counterpart to Top-$k$ sampling, Small-$k$ sampling selects the $k$ samples with the lowest density scores. 

As shown in Figure~\ref{fig:top_random_small}, to validate the Top-k strategy inversely, we observe that sampling low-density regions leads to a decrease in IS while simultaneously causing FID to rise due to the density guided search mechanism.

\subsubsection{Random-$k$ Sampling.} 
We randomly select $k$ samples from the sampling space $\mathcal{X}$. Then, based on the density scores of these $k$ samples, we choose one final sample:

\begin{equation}
S_{\text{random-}k} = \{ x_i \mid x_i \in \mathcal{X},\ i = 1, 2, \ldots, k \},
\end{equation}

\begin{equation}    
\mathrm{Random}_{k}(m, \mathcal{X}) := 
\left\{
\begin{aligned} 
\mathop{\mathrm{sample}}\limits_{|\kappa|=k,\,\kappa \subset \mathcal{X}} \mathcal{P}(\kappa), & \text{if } |\mathcal{X}| > k, \\
\mathcal{X}, & \text{otherwise},
\end{aligned}
\right.
\end{equation}
where $\mathcal{P}(\kappa)$ is the probability distribution over subsets $\kappa \subset \mathcal{X}$. The probability of selecting each sample $x_i$ is normalized.

However, as shown in Figure~\ref{fig:top_random_small}, with inference time scaling, although FID remains stable, the improvement in IS is limited. This is due to the lack of a guiding mechanism for finding representative samples.

\begin{figure*}[t]
    \centering
    \includegraphics[width=1\textwidth]{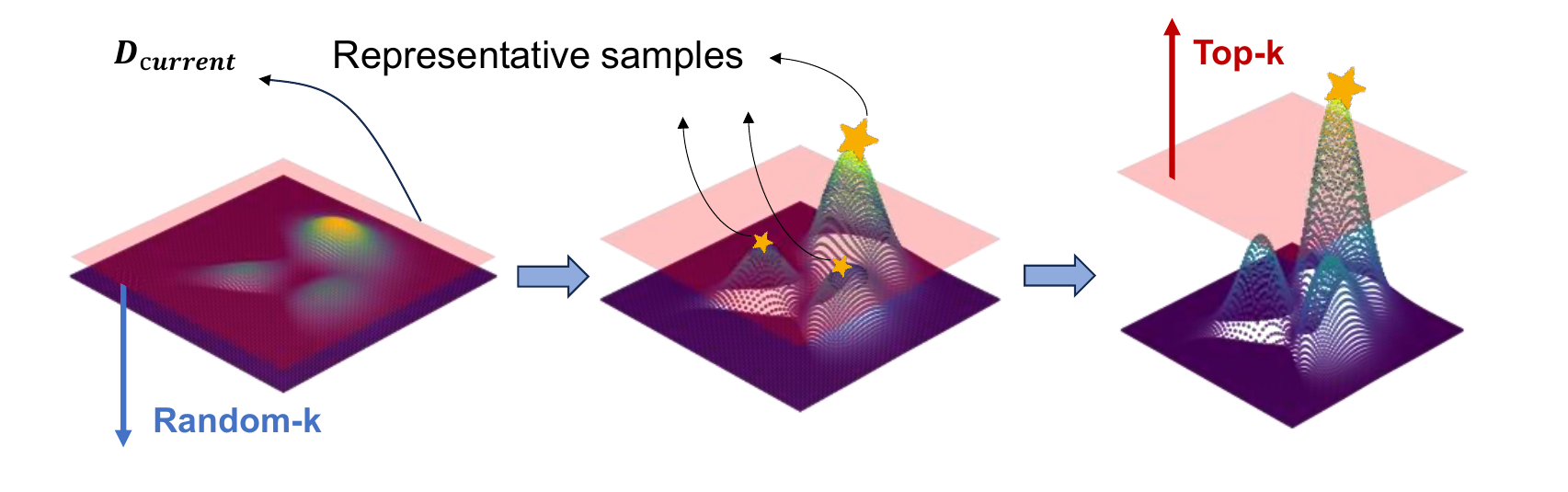}
    \caption{Across successive expansion iterations (left → right), the representative samples of distribution exhibits progressive structural refinement.}
    \label{fig:density}
    \vspace{-10pt}
\end{figure*}

\subsubsection{Density-guided representative samples (VAR-Scaling).}
To balance the diversity of categories and the quality of images, our examination of Section~\ref{subsec:inference_scaling} shows that the characteristics of the subspace fall into two categories:
\emph{i)}. High-density regions: Samples in these regions may correspond to archetypal, stable patterns intensively learned from training data, characterized by highly concentrated distribution.
\emph{ii)}. Low-density regions: Samples in these regions may represent marginal cases or outliers within the model's predictive distribution, characterized by highly scattered distribution.

Based on these findings, we adopt different sampling strategies:
\emph{i)}. For high-density regions, we apply top-k sampling.
\emph{ii)}. For low-density regions, we use random-k sampling.
As show in Figure~\ref{fig:density}, the details on how to distinguish between high and low density are as follows:

    \begin{equation}
   \begin{cases} 
   \text{Density Mean} = \frac{1}{n} \sum_{i=1}^{n} \hat{f}(x_i)\\
   D_{\text{current}} = \text{Density Mean} \times \alpha
   \end{cases}
   \end{equation}
   where \( n \) is the number of samples, and \( \hat{f}(x_i) \) is the density score of sample \( x_i \). $\alpha$ is the density threshold coefficient.
   Finally, we can get: 
   \[
   \text{Density Classification} = 
   \begin{cases} 
   \text{low-density} & \text{if } D_{\text{current}} < D_{\text{max}} \\
   \text{high-density} & \text{otherwise}
   \end{cases}
   \]

\section{Experiments}

\subsection{Settings}
\subsubsection{Datasets and models.} For VAR~\cite{tian2024visual} and FlexVAR~\cite{jiao2025flexvar}, we conduct experiments on the ImageNet~\cite{deng2009imagenet} 256×256 conditional generation benchmark such as Fréchet inception distance (FID~\cite{heusel2017gans}), inception score (IS~\cite{salimans2016improved}). For evaluation of Infinity models, we report results on text-to-image benchmarks like GenEval~\cite{ghosh2023geneval}.

\input{table/implementation}
\vspace{-10pt}
\subsubsection{Implementation details.}
The comprehensive standard experimental configurations for all models are meticulously listed in Table~\ref{tbl:main}. This table explicitly specifies the default settings consistently used throughout the experiments.

\subsection{Image Generation}

\subsubsection{Evaluations on class-conditional and text-conditional image generation benchmarks.} 
\begin{figure*}[htp]
    \centering
    \begin{subfigure}[b]{0.48\textwidth}
        \centering
        \includegraphics[width=\linewidth]{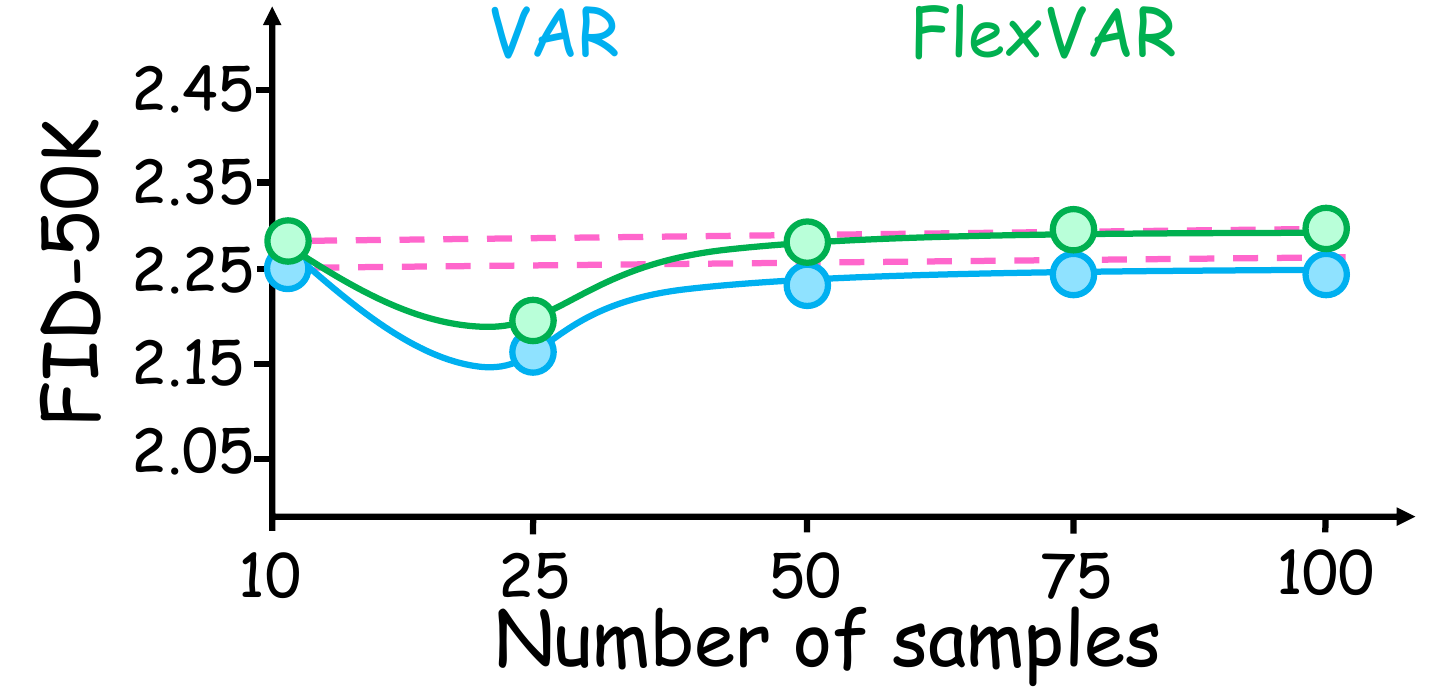}
        \caption{FID $\downarrow$ evaluated on ImageNet-50k}
        \label{fig:fid}
    \end{subfigure}
    \hfill
    \begin{subfigure}[b]{0.48\textwidth}
        \centering
        \includegraphics[width=\linewidth]{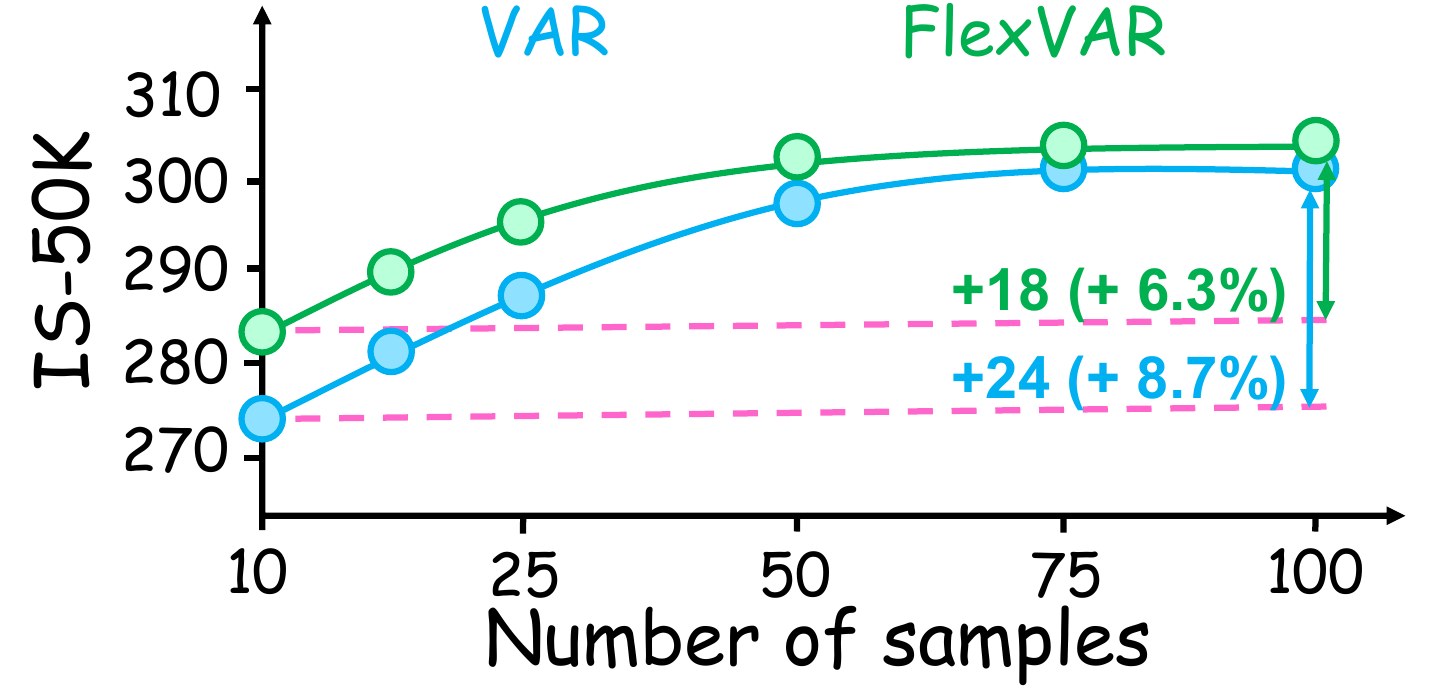}
        \caption{IS $\uparrow$ evaluated on ImageNet-50k}
        \label{fig:is}
    \end{subfigure}
    \vspace{-5pt}
    \caption{Experimental results demonstrate that high-density regions correspond to representative samples, whereas low-density samples are low-quality samples.}
    \label{fig:test}
\end{figure*}

As shown in Figures~\ref{fig:test} and~\ref{fig:infinity}, VAR-Scaling achieves notable inference gains in both class-conditional (VAR~\cite{tian2024visual}, FlexVAR~\cite{jiao2025flexvar}) and text-to-image (Infinity~\cite{han2025infinity}) tasks: +8.7\% IS~\cite{salimans2016improved}, +6.3\% IS, and +1.1\% Geneval~\cite{ghosh2023geneval}.

\begin{minipage}{0.45\textwidth}
\setlength{\leftskip}{-0.5cm}
\subsubsection{Analyses.} 
Given that VAR and FlexVAR employ VQ while Infinity utilizes binary spherical quantization (BSQ~\cite{zhao2024image}), we accordingly configure distinct search times to construct appropriately-sized sampling spaces for identifying optimal representative samples.
\end{minipage}%
\hfill
\begin{minipage}{0.48\textwidth}
\centering
\includegraphics[width=\linewidth]{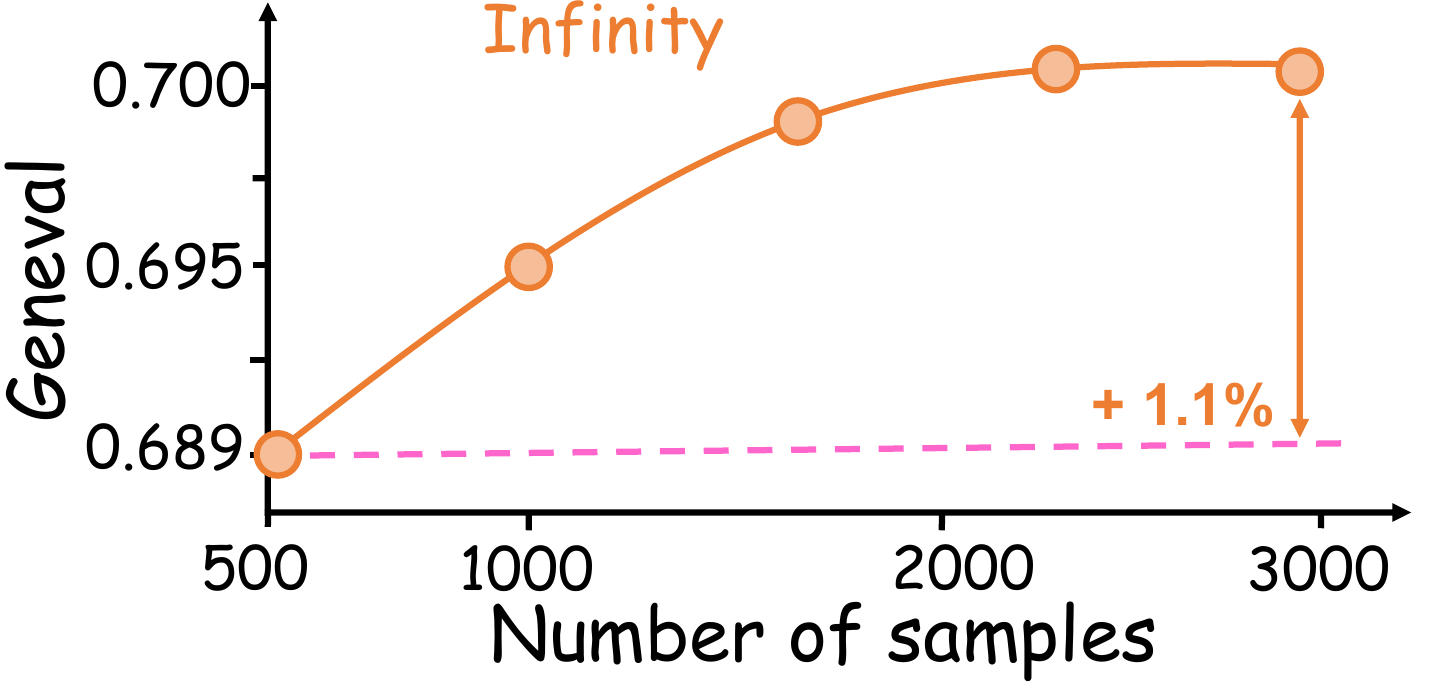}
\vspace{-20pt}
\captionof{figure}{Geneval $\uparrow$ Evaluation of Infinity}
\label{fig:infinity}
\end{minipage}

\subsection{Ablation Study}
In this section, we carry out a series of ablation studies to explore various interesting features of VAR-Scaling in different scenarios. Unless noted, all experiments follow the default configuration and report mean performance.

\vspace{-10pt}
\input{table/aba_scales}

\input{table/aba_all}
\vspace{-30pt}
\subsubsection{Ablation on scales.} Aimed at identifying the most critical scale for sampling, we conducted multi-scale experiments on general patterns. Setting: We change the default scale. Detaildly, we select the d30 model of VAR, use L2 distance as the distance metric, set the cfg weight to 1.0, and follow the sampling number, coreset size, and $\alpha$ settings shown in Table.~\ref{tbl:main}. We conduct experiments at each scale and compare the FID~\cite{heusel2017gans} and IS~\cite{salimans2016improved} results. Based on Table.~\ref{tbl:exp_abl_scales} it is clear that each scale has a different weight. When scale = 0, tuning the hyperparameters also struggles to control FID, as VAR-Scaling introduces biased information too early, causing the search to converge prematurely to the verifier's bias. For scales $\in \{2,3,4\}$, their effects are very similar, but they require more computational resources compared to scale = 1. Scales $\in \{5,6,7,8,9\}$ are specific patterns that mainly affect local details, resulting in a smaller improvement. Overall, through numerous experiments, we explained that the importance and effects of the VAR~\cite{tian2024visual} scales vary, and we selected scale = 1 as the optimal processing layer.

\subsubsection{Ablation on model sizes.}

To investigate the generalization of VAR-Scaling under different model sizes, we conducted experiments. Setting:We change the default model size and density threshold coefficient. As shown in Table.~\ref{tbl:exp_abl_model_size}, experiments demonstrate that our method generalizes across different model sizes, and we have found that the optimal configuration is constantly changing, with no universal solution.

\subsubsection{Ablation on different strategies.} 

VAR-Scaling's efficacy is validated through comparisons with top-k and random-k methods. Table~\ref{tbl:exp_abl_α} shows top-k improves IS~\cite{salimans2016improved} over random-k but misaligns with FID~\cite{heusel2017gans} objectives by prioritizing single-sample quality over population diversity. This diversity deficit reduces sample variance, inducing mode collapse at larger sampling sizes. Random-k's unconstrained search yields marginal gains. VAR-Scaling successfully integrates both approaches' strengths, confirming our design rationale.

\subsubsection{Ablation on the number of representative samples.} 

To explore the impact of different number of representative samples on VAR-Scaling, we conducted experiments. Setting: we vary the number of representative samples. As shown in Table.~\ref{tbl:exp_abl_coreset_size}, experiments demonstrate that our method is insensitive to the size of the core set. This means that the samples selected by VAR-scaling are representative, and we have proven that the representative samples are indeed the central samples.

\subsubsection{Ablation on classifier-free guidance.} 

To investigate the generalization of VAR-Scaling under different classifier-free guidance weights, we conducted experiments. Setting: We change the default classifier-free guidance $\in \{1.0, 1.5, 2.0\}$. As show in Table.~\ref{tbl:exp_abl_cfg}, the study found that our method is effective not only when guidance is not used, but also under strong classifier-free guidance weight influence, where our method can further enhance the performance. This indicates that our method generalizes across different classifier-free guidance.

\subsubsection{Ablation on distance metric.} 
Comparisons of Euclidean distance and cosine similarity reveal key robustness properties. As shown in Figure~\ref{tbl:exp_abl_mertic}, the marginal differences in FID/IS metrics further demonstrate VAR-Scaling's strong flexibility across diverse measurement choices.

\subsection{Visualization}
\begin{figure*}[h]
    \centering
    \includegraphics[width=1\textwidth]{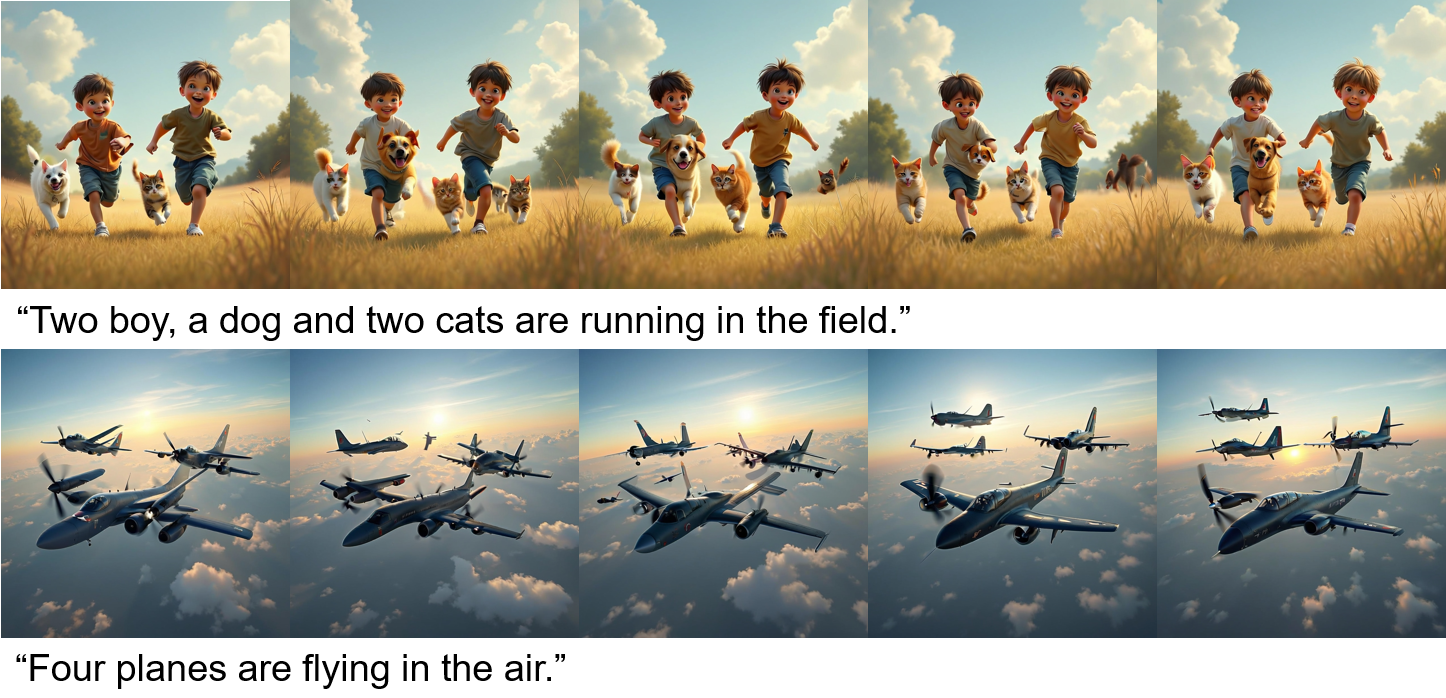}
    \caption{The visualization of VAR-Scaling's segmentation results on Infinity. As the scaling increases, the quality of the image becomes more stable and is better able to focus on the information provided by the prompt.}
    \label{fig:vis}
\end{figure*}

In this section, we present abundant visualization of VAR-Scaling’s outputs on Infinity~\cite{han2025infinity}.
Figure~\ref{fig:vis} demonstrates that in inference-time scaling process, the generated outputs exhibit increasing alignment with prompt specifications, particularly for quantitative evaluation tasks such as counting.

\section{Conclusion and Future Work}

In this work, we propose the first framework for inference time scaling in VAR. 
It provides a new approach for the inference time scaling of discrete spaces for the first time. 
Extensive experiments have supported our method's effectiveness, and a series of ablation studies also validates its robustness in data and parameters.
However, our method does have some limitations:
\emph{i)}. The theoretical linkage between high-density samples and high-quality representative samples warrants deeper investigation.
\emph{ii)}. VAR-Scaling has not yet been applied to video generation models.
Conclusively, our method is an effective solution for VAR, and we hope to inspire future researches in generative model optimization.

\subsubsection{\ackname} This research was funded in part by the National Natural Science Foundation of China under Grant 62372355, in part by the Natural Science Basic Research Program of Shaanxi Province under Grant 2023-JC-ZD-39 and 2024JC-YBMS-520, in part by the Open Fund of State Key Laboratory of Intelligent Manufacturing of Advanced Construction Machinery(KFKT2024008).

{\let\clearpage\relax  

\bibliography{references}
} 

\end{document}

%% file: commands.tex
\newcommand{\tablestyle}[2]{\setlength{\tabcolsep}{#1}\renewcommand{\arraystretch}{#2}\centering\footnotesize}
\renewcommand{\paragraph}[1]{\vspace{1.25mm}\noindent\textbf{#1}}

\newcolumntype{x}[1]{>{\centering\arraybackslash}p{#1pt}}
\newcolumntype{y}[1]{>{\raggedright\arraybackslash}p{#1pt}}
\newcolumntype{z}[1]{>{\raggedleft\arraybackslash}p{#1pt}}

\newcommand{\app}{\raise.17ex\hbox{$\scriptstyle\sim$}}

\definecolor{deemph}{gray}{0.6}

\definecolor{baselinecolor}{gray}{.9}

\definecolor{dt}{HTML}{ADCAD8}
\definecolor{dt2}{HTML}{cddfe7}

\definecolor{defaultcolor}{HTML}{E8E2F7}
\newcommand{\default}[1]{\cellcolor{defaultcolor}{#1}}

\let\cite\citep

\renewcommand{\paragraph}[1]{\vspace{1.25mm}\noindent\textbf{#1}}
\newlength\savewidth

\newcolumntype{x}[1]{>{\centering\arraybackslash}p{#1pt}}
\newcolumntype{y}[1]{>{\raggedright\arraybackslash}p{#1pt}}
\newcolumntype{z}[1]{>{\raggedleft\arraybackslash}p{#1pt}}
\definecolor{degray}{gray}{.6}

%% file: table/implementation.tex
\begin{table*}[htp]
\centering
\setlength{\tabcolsep}{2pt}
\caption{Default configuration and recommended parameter range.}
\vspace{-5pt}
\label{tbl:main}

\scriptsize

\begin{subtable}{\linewidth}
    \centering
    \tiny
    \tablestyle{9.0pt}{1.0}
        \caption{The default configurations of the three models.}
    \label{tbl:implementation}
\begin{tabular}{lccc}
\hline
default & VAR~\cite{tian2024visual} & FlexVAR~\cite{jiao2025flexvar} & Infinity~\cite{han2025infinity} \\ 
\hline
GPU  & A6000 48GB & A6000 48GB & A100 40GB \\
scale & 1 & 2 & 3 \\
model size & 16/20/24/30 & 24 & 2B  \\
number of samples (\#n) & 10-100 & 10-100 & 500-3000 \\
Classifier-Free Guidance (cfg) & 1.0 & 2.5 & 4.0 \\
representative samples (\#rep) & 10 & 10 & 500 \\
\hline
\end{tabular}
\end{subtable}

\begin{subtable}{\linewidth}
    \centering
    \tiny
    \tablestyle{5.6pt}{1.0}
    \caption{density threshold coefficient range recommended.}
    \label{tbl:exp_settings_alpha}

    \begin{tabular}{ccccccc}
    \hline
    \multicolumn{1}{c}{\centering default} & \multicolumn{4}{c}{\centering VAR~\cite{tian2024visual}} & \multicolumn{1}{c}{\centering Flex-VAR~\cite{jiao2025flexvar}} & \multicolumn{1}{c}{\centering Infinity~\cite{han2025infinity}} 
    \multirow{3}{*}{}
    \\
    \hline
    size
    &
    d16
    &
    d20
    &
    d24
    &
    \default{d30}
    &
    \default{d24}
    &
    \default{2B}
    \\
    
    $\alpha$
    &
    2.3 - 2.9
    &
    2.2 - 2.8
    &
    2.1 - 2.7
    &
    \default{2.0 - 2.6}
    &
    \default{2.5 - 2.8}
    &
    \default{1.0}
    \\
    \hline
    \end{tabular}
\end{subtable}
\end{table*}

%% file: table/aba_scales.tex
\begin{table}[H]
    \centering
    \tablestyle{5.5pt}{1.0}
    \begin{tabular}{ccccccc}
    \multicolumn{6}{c}{\centering VAR} \\
     scale& [H,W] &number of samples & representative samples &$\alpha$ & FID $\downarrow$ & IS $\uparrow$\\
    \hline
    -&- & - & - & - & 2.25 & 274.7 \\
    \hline
    \multirow{3}{*}{0} &[1,1] &20&10&2.3 & 2.44 & 285.2 \\
     &[1,1]& 50&10&2.4 & 3.07 & 288.2 \\
     &[1,1]& 100&10&  2.5 & 2.91 & 296.0 \\
    \hline
    

    \rowcolor{defaultcolor}
    &[2,2] &20 &10 & 2.3 &2.16 &279.3 \\
    \rowcolor{defaultcolor}
    1&[2,2] &50 &10 &2.4 &2.25 &298.7 \\
    \rowcolor{defaultcolor}
    &[2,2] &100 &10 &2.4 &2.19 &295.2 \\
    
     \hline
     \multirow{3}{*}{2} &[3,3]& 20&10&1.7 & 2.18 & 283.3 \\
     &[3,3]& 50&10&1.7 & 2.25 & 298.3 \\
     &[3,3]& 100&10&1.8 & 2.24 & 296.2 \\
      \hline
          \multirow{3}{*}{3} &[4,4]& 20&10&1.9 & 2.17 & 283.2 \\
     &[4,4]& 50&10&1.9 & 2.23 & 289.4 \\
     &[4,4]& 100&10&1.8 & 2.33 & 291.1 \\
      \hline
          \multirow{3}{*}{4} &[5,5]& 20&10&1.9 & 2.19 & 288.2 \\
     &[5,5]& 50&10&1.9 & 2.21 &  295.8\\
     &[5,5]& 100&10&2.0 & 2.12 & 288.3 \\
      \hline
          \multirow{3}{*}{5} &[6,6]& 20&10&1.8 & 2.21 & 289.0 \\
     &[6,6]& 50&10&1.8 & 2.25 & 293.0 \\
     &[6,6]& 100&10&2.0 & 2.10 & 285.4 \\
      \hline
          \multirow{3}{*}{6} &[8,8]& 20&10&1.7 & 2.10 & 284.3 \\
     &[8,8]& 50&10&1.7 &2.26 & 288.2 \\
     &[8,8]& 100&10&1.9 & 2.09 & 285.4 \\
      \hline
          \multirow{3}{*}{7} &[10,10]& 20&10&1.4 & 2.18 & 286.5 \\
     &[10,10]& 50&10&1.4 & 2.20 & 287.1 \\
     &[10,10]& 100&10& 1.5 & 2.11 & 284.7 \\
      \hline
    \multirow{3}{*}{8} &[13,13]& 20&10&1.2 & 2.18 & 282.7 \\
     &[13,13]& 50&10& 1.2 & 2.24 & 286.2 \\
     &[13,13]& 100&10&1.3 & 2.12 & 284.5 \\
      \hline

    \multirow{3}{*}{9} &[16,16]& 20&10&1.1 & 2.19 & 280.2 \\
     &[16,16]& 50&10&1.1 & 2.23 & 284.3 \\
     &[16,16]& 100&10&1.3 & 2.16 & 281.8 \\
      \hline

    \end{tabular}
    \caption{Ablation on scales $\in \{0,1,2,3,4,5,6,7,8,9\}$
 and all scales $\ge$ 1. Scaled 1 is selected, considering the cost of high scales and performance.}
    \label{tbl:exp_abl_scales}
\end{table}

%% file: table/aba_all.tex
\begin{table*}[htp]
\centering
\setlength{\tabcolsep}{2pt}
\vspace{-1.0em}
\caption{Ablation experiment result chart.}
\label{tab:main}
\vspace{1em}
\scriptsize
\vspace{-10pt}
\begin{subtable}{\linewidth}
    \centering
    \tiny
    \tablestyle{7.2pt}{0.9}
    \caption{Ablation study on model sizes. VAR~\cite{tian2024visual} d30 is selected, considering it better demonstrates the effectiveness of the method.}
    \label{tbl:exp_abl_model_size}
    \begin{tabular}{cccccc}
    \hline
     model & number of samples & representative samples &$\alpha$ & FID $\downarrow$ & IS $\uparrow$\\
    \hline

    \multirow{4}{*}{VAR-d16} 
     & - & -&-& 3.81 & 227.2 \\
     & 20&10&2.4 & 3.77 & 227.5 \\
     & 50&10&2.4 & 3.75 & 228.3 \\
     & 100&10&2.6 &  3.69& 232.6 \\
    \hline
    \multirow{4}{*}{VAR-d20} 
     & - &- &-& 2.84& 253.3 \\
     & 20&10&2.3 & 2.72 & 267.4 \\
     & 50&10&2.4 & 2.81 & 262.3 \\
     & 100&10&2.5 & 2.78 & 261.0\\      
     \hline
     \multirow{4}{*}{VAR-d24} 
     & - & -& -& 2.28 & 273.7 \\
     & 20&10&2.7 & 2.18 & 275.2 \\
     & 50&10&2.7 & 2.24 & 276.3 \\
     & 100&10&2.8 &  2.21 & 280.3 \\
      \hline
    \rowcolor{defaultcolor}
     & - &-&-& 2.25 & 274.7 \\
     \rowcolor{defaultcolor}
     & 20&10&2.3 & 2.16 & 279.3 \\
     
    \rowcolor{defaultcolor}
& 50 & 10 & 2.4 & 2.25 & 298.7 \\
\rowcolor{defaultcolor}
\multirow{-4}{*}{VAR-d30}
& 100 & 10 & 2.4 & 2.19 & 295.2 \\
\hline
\end{tabular}
\end{subtable}

\begin{subtable}{0.48\linewidth}
    \centering
    \tiny
    \tablestyle{2pt}{0.9}
        \caption{Ablation study on different strategies. Selected VAR-Scaling, considering performance.}
    \label{tbl:exp_abl_α}
    
        \begin{tabular}{cccccc}
        \hline
         strategy & \#n & \#rep & $\alpha$ & FID $\downarrow$ & IS $\uparrow$ \\
        \hline
         -& -&- & - & 2.25 & 274.7 \\
        \hline
          \rowcolor{defaultcolor}
        & 20&10&2.3 & 2.16 & 279.3 \\
          \rowcolor{defaultcolor}
         & 50&10&2.4 & 2.25 & 298.7 \\
           \rowcolor{defaultcolor}
           \multirow{-3}{*}{VAR-Scaling} 
         & 100&10&2.4 & 2.19 & 295.2 \\
         \hline
          \multirow{3}{*}{top-k}  & 20&10& -& 2.30 & 296.9 \\
         & 50&10& -& 2.46 & 304.7 \\
         & 100&10& -& 2.42 & 301.5 \\
          \hline
         \multirow{3}{*}{random-k}  & 20&10&- & 2.25 & 276.1 \\
         & 50&10 & -& 2.27&  276.7\\ 
         & 100&10 &- & 2.24&  275.3\\
         \hline
        \end{tabular}
\end{subtable}
\hfill
\begin{subtable}{0.48\linewidth}
    \centering
    \tiny
    \tablestyle{7.6pt}{0.9}
        \caption{Ablation study on coreset size. A size of 10 is selected, balancing model performance and computational cost.}
    \label{tbl:exp_abl_coreset_size}
\begin{tabular}{ccccc}
\hline
 \#n & \#rep & $\alpha$& FID $\downarrow$ & IS $\uparrow$\\
\hline
 -& -&- & 2.25 & 274.7 \\
\hline
\multirow{3}{*}{5} 
 & 20&2.3 & 2.17 & 279.0 \\
 & 50&2.4 & 2.23 & 297.5 \\
 & 100&2.4 & 2.21 & 295.1 \\
\hline
\rowcolor{defaultcolor}
  & 20&2.3 & 2.16 & 279.3 \\
  
  \rowcolor{defaultcolor}
 & 50&2.4 & 2.25 & 298.7 \\
  \rowcolor{defaultcolor}
 \multirow{-3}{*}{10}
 & 100&2.4 & 2.19 & 295.2 \\
 \hline
\multirow{3}{*}{15}  & 20&2.3 & 2.13 & 278.8 \\
 & 50&2.4 & 2.24 & 298.4 \\
 & 100&2.4 & 2.17 & 294.0 \\
 \hline
\end{tabular}
\end{subtable}

\begin{subtable}{0.48\linewidth}
    \centering
    \tiny
    \caption{Ablation study on cfg $\in \{1.0,1.5,2.0\}$. cfg=1.0 is selected to minimize the influence of cfg and demonstrate the capability of VAR-Scaling in the base model, focusing on the simple conditional generation task without guidance.}
    \label{tbl:exp_abl_cfg}
    \tablestyle{3.8pt}{0.9}
    \begin{tabular}{cccccc}
     \\
     \hline
     cfg & \#n &\#rep &$\alpha$ & FID $\downarrow$ & IS $\uparrow$\\
    \hline
  \rowcolor{defaultcolor}
     & - &- &-&2.25 & 274.7 \\
       \rowcolor{defaultcolor}
     & 20&10&2.3 & 2.16 & 279.3 \\
       \rowcolor{defaultcolor}
     & 50&10&2.4 & 2.25 & 298.7 \\
       \rowcolor{defaultcolor}
     \multirow{-4}{*}{cfg=1.0} 
     & 100&10&2.4 & 2.19 & 295.2 \\
    \hline
    \multirow{4}{*}{cfg=1.5} 
     & - &-&-&2.01 & 305.7 \\
     & 20&10&1.8 &1.97 & 310.4\\
     & 50&10&1.8 & 2.16 &  320.4\\
     & 100&10&1.9 & 2.13 & 319.1 \\
     \hline
    \end{tabular}
\end{subtable}
\hfill
\begin{subtable}{0.48\linewidth}
    \centering
    \tiny
    
    \caption{Ablation study on distance metrics. ED: Euclidean Distance, CS: Cosine Similarity. Euclidean Distance is selected as it more effectively captures scaling patterns during inference, particularly showcasing consistent performance gains across varying model configurations.} 
    \label{tbl:exp_abl_mertic}
    \tablestyle{4.1pt}{0.9}
    \begin{tabular}{cccccc}
    \hline
    metric & \#n &\#rep&$\alpha$ & FID $\downarrow$ & IS $\uparrow$ \\
    \hline
    - & - &-&-& 2.25 & 274.7 \\
    \hline
    
     \rowcolor{defaultcolor}
    & 20&10&2.3 & 2.16 & 279.3 \\
     \rowcolor{defaultcolor}
     & 50&10&2.4 & 2.25 & 298.7 \\
      \rowcolor{defaultcolor}
      \multirow{-3}{*}{ED} 
     & 100&10&2.4 & 2.19 & 295.2 \\
     \hline
    \multirow{3}{*}{CS} & 20&10&2.3 & 2.23 & 296.8 \\
     & 50&10&2.4 & 2.21 & 293.0 \\
     & 100&10&2.4 & 2.18 & 288.1 \\
     \hline
    \end{tabular}
    
\end{subtable}

\end{table*}